\documentclass[journal]{IEEEtran}
\usepackage[square,sort,comma,numbers]{natbib}

\ifCLASSINFOpdf
\else
\fi
\usepackage{amsmath}
\usepackage{amssymb}
  {

  }

\usepackage{hhline}
\usepackage[hyphens]{url}  
\usepackage{subcaption}
\usepackage{tcolorbox}
\usepackage{color}
\usepackage{colortbl}
\usepackage[makeroom]{cancel}
\definecolor{myred}{rgb}{0.8,0,0}
\definecolor{mygreen}{rgb}{0,0.6,0}
\definecolor{myblue}{rgb}{0,0,0.7}

\definecolor{DarkGray}{gray}{0.9}
\definecolor{MediumGray}{gray}{0.75}
\definecolor{LightGray}{gray}{0.5}

\usepackage{algorithm}
\usepackage[noend]{algpseudocode}

\def\bbbr{{\rm I\!R}}

\newcommand{\R}{{\bbbr}{}}

\usepackage{array}
\newcolumntype{P}[1]{>{\centering\arraybackslash}p{#1}}
\newcolumntype{M}[1]{>{\centering\arraybackslash}m{#1}}
\begin{document}
%
\title{Single-Reset Divide \& Conquer Imitation Learning}
%
%
%

\author{Alexandre~Chenu$^{1}$,
        Olivier~Serris$^{1}$,
        Olivier~Sigaud$^{1}$,
        and~Nicolas~Perrin-Gilbert$^{1}$
\thanks{$^{1}$Sorbonne Université, CNRS, Institut des Systèmes Intelligents et de Robotique, ISIR F-75005 Paris, France. \textit{\{\mbox{chenu}, \mbox{serris}, \mbox{sigaud}, \mbox{perrin}\}@isir.upmc.fr}. Corresponding author: Alexandre Chenu.}
}

\maketitle

\begin{abstract}

Demonstrations are commonly used to speed up the learning process of deep reinforcement learning algorithms. 
To cope with the difficulty of accessing multiple demonstrations, some algorithms have been developed to learn from a single demonstration. 
In particular, the Divide \& Conquer Imitation Learning algorithms leverage a sequential bias to learn a control policy for complex robotic tasks using a single state-based demonstration. 
The latest version, DCIL-II demonstrates remarkable sample efficiency. 
This novel method operates within an extended Goal-Conditioned Reinforcement Learning framework, ensuring compatibility between intermediate and subsequent goals extracted from the demonstration. 
However, a fundamental limitation arises from the assumption that the system can be reset to specific states along the demonstrated trajectory, confining the application to simulated systems. 
In response, we introduce an extension called Single-Reset DCIL (SR-DCIL), designed to overcome this constraint by relying on a single initial state reset rather than sequential resets. 
To address this more challenging setting, we integrate two mechanisms inspired by the Learning from Demonstrations literature, including a Demo-Buffer and Value Cloning to guide the agent toward compatible success states. 
In addition, we introduce Approximate Goal Switching to facilitate training to reach goals distant from the reset state. 
Our paper makes several contributions, highlighting the importance of the reset assumption in DCIL-II, presenting the mechanisms of SR-DCIL variants and evaluating their performance in challenging robotic tasks compared to DCIL-II. 
In summary, this work offers insights into the significance of reset assumptions in the framework of DCIL and proposes SR-DCIL, a first step toward a versatile algorithm capable of learning control policies under a weaker reset assumption.

\end{abstract}


%
\IEEEpeerreviewmaketitle

\section{Introduction}

The Divide \& Conquer Imitation Learning algorithms \citep{Chenu2022divide, chenu2022leveraging} are novel Deep Reinforcement Learning (DRL) algorithms that leverage a sequential bias to learn a control policy for complex robotic tasks using a single demonstration. 
This approach can be used to learn a goal-conditioned policy to control the system between successive intermediate low-dimensional goals. It is based on an extended Goal-Conditioned RL (GCRL) framework designed to ensure that the state resulting from reaching an intermediate goal is compatible with the achievement of the following goal. 
Although the approach shows unprecedented sample efficiency when applied to complex robotics tasks such as grasping or humanoid locomotion, it relies on a strong assumption that the system can be reset to some states selected in the demonstrated trajectory. 
This assumption limits the approach to simulated systems. Indeed, when tackling a complex robotic task such as locomotion with an under-actuated humanoid robot, one can easily imagine how difficult it would be to reset the robot to a demonstrated state corresponding to a complex configuration or including precise angular and positional velocities.  

In this paper, we propose an extension of the DCIL-II algorithm \cite{chenu2022leveraging} called Single-Reset DCIL (SR-DCIL). 
As DCIL-II, SR-DCIL learns a goal-conditioned policy to control the system between a sequence of low-dimensional goals.
In contrast, SR-DCIL only assumes that the robot can be reset to a single state at the beginning of the demonstration.
To adapt DCIL-II to this more challenging setting, we call upon mechanisms inspired by the literature on Learning from Demonstrations \citep{atkeson1997robot} that are likely to be compatible with the use of a single demonstration. From these mechanisms, we extract a Demo-Buffer (DB) and a Value Cloning (VC) mechanism. 
DB reuses demonstration transitions during training to increase the value of associated state-action pairs. Unlike DCIL-II, the actions of the demonstration must be available to use the DB, which corresponds to an additional assumption.
On the other hand, VC does not assume the availability of these actions. It estimates the theoretical value of each state in the demonstration and forces the learned value function to match this theoretical value using regression. 
Both mechanisms are designed to guide the agent towards states compatible with the achievement of future goals, while reaching intermediate ones.
In addition, we introduce a third mechanism: Approximate Goal Switching (AGS). Based on a dynamic threshold of the Q-value function, this mechanism allows the agent to change from one intermediate goal to the next, even if the current goal has not been reached exactly. It is conceived to help the agent train for goals distant from the reset state. 

This paper is organized as follows. First, in section~\ref{sec:related_sr_dcil}, we cover related work stemming from GCRL and RL from demonstration. Section~\ref{sec:background_dcil} provides background on the DCIL framework.
In Section~\ref{sec:methods_sr_dcil} we first highlight the capital importance of the reset hypothesis in DCIL-II. Then, to tackle the limitations induced by the weaker assumption of a system that can only be reset to a single state, we present three mechanisms (DB, VC and AGS) that can be combined with DCIL-II to obtain the different variants of the SR-DCIL algorithm.
In Section~\ref{sec:SR_DCIL_xp}, we evaluate these variants in two challenging robotics tasks and compare it to DCIL-II. 
The first experiment validates the three mechanisms in a low-dimensional non-holonomic navigation task. The second highlights the difficulty of SR-DCIL in scaling up to complex high-dimensional environments. 
In summary, our main contributions are: (1) we highlight the importance of the strong reset assumption made by DCIL-II; (2) under the name SR-DCIL, we propose several variants of DCIL-II capable of learning a control policy under a weaker reset assumption.


\section{Related Work}
\label{sec:related_sr_dcil}

As an extension of DCIL-II \cite{chenu2022leveraging}, SR-DCIL extracts a sequence of goals from a demonstration and learns to reach them sequentially to reproduce the complex demonstrated behavior. As explained in Section~\ref{sec:seq_goal_reach}, this strategy is adopted by various classes of RL algorithms.
In addition, to relax the reset assumption of DCIL, SR-DCIL uses mechanisms inspired from the literature RL from Demonstration which is presented in Section~\ref{sec:guiding_rl_demo}. 

\subsection{Sequential Goal Reaching}
\label{sec:seq_goal_reach}
To achieve a distant goal, DCIL algorithms rely on sequential goal reaching, a strategy used in different classes of RL algorithms. 
For instance, in Hierarchical RL, a high-level policy may construct a sequence of goals which must be successively achieved by a low-level goal-conditioned policy  \citep{dayan1992feudal, nachum2018data, Levy2019LearningMH, li2022hierarchical, bagaria2019option, bagaria2021robustly}. 
Other approaches combine goal-conditioned value learning and goal-level planning to achieve distant goals  \citep{nair2018visual, nasiriany2019planning, eysenbach2019search, zhang2021c, chane2021goal}.
However, in order to exploit an accurate value function, it should be noted that these planning methods are based on at least one of the two following assumptions: a dense and informative reward function is available, or an assumption of resetting in hard-to-attain states is required. two hypotheses that are not valid in the problems tackled by SR-DCIL.

Depending on the nature of the goals, it may be difficult to chain several goals sequentially. It is therefore important to reach each goal by passing through states from which it is possible to reach the following ones, or to select a specific goal sequence that the agent will be able to chain together. 
Value propagation mechanisms \cite{Chenu2022divide, chenu2022leveraging, bagaria2021robustly} can be used to achieve both these objectives. 
In DCIL, this mechanism requires a strong reset assumption. To relax this hypohesis, SR-DCIL combines the value propagation mechanism of DCIL with techniques inspired by the RL from demonstration literature.

\subsection{Guiding RL using demonstrations}
\label{sec:guiding_rl_demo}
RL and IL can be merged in different ways: offline RL \citep{prudencio2022survey}, inverse RL \citep{arora2021survey}, RL from demonstration (RLfD) \citep{schaal1996learning}.
However, only a few mechanisms inspired by the literature of RLfD can be used in a complementary manner with another RL algorithm.
Indeed, on the one hand, offline RL learns directly from an offline dataset of interactions and assumes no additional interaction with the environment. 
On the other hand, inverse RL uses the demonstration to learn a reward function that may interfere with an already available one. 
On the contrary, in RLfD, one simply uses the demonstration along with additional training interactions collected by any RL agent. 

Using the demonstration in RLfD can help the agent to learn more effectively at different levels. 
If both the demonstrated states and actions are available, the latter can guide the policy in the action selection process. 
Alternatively, demonstrations can be used to help the critic estimate the value function.

In the SACR2 paper \citep{martin2021learning}, the authors list several variants of the Soft Actor-Critic (SAC) algorithm \citep{haarnoja2018soft} augmented with different mechanisms inspired from the RLfD literature. 

First, SAC Behavioral Cloning (SACBC) uses Behavioral Cloning \citep{pomerleau1991efficient} as a regularization mechanism for the update of the actor. This is inspired by various other RLfD algorithms \citep{nair2018overcoming, goecks2019integrating, fujimoto2021minimalist}. These methods all propose two main components. First, a secondary replay buffer is filled with transitions from demonstrations. Using these transitions, an auxiliary BC loss is computed and added to the original policy loss. 
However, BC usually performs poorly when only a single demonstration is available \citep{pmlr-v9-ross10a, resnick2018backplay, behbahani2019learning}. 
Therefore, this mechanism is not relevant to the extremely weak data regime where only a single demonstration is available. 

In SAC from Demonstrations (SACfD), which is inspired by \citep{vecerik2017leveraging, paine2019making}, a similar secondary replay buffer is filled with transitions from demonstrations. However, instead of using those transitions to compute an additional BC loss, they are directly used as training transitions during the critic update. 
Similarly, Soft Q Imitation Learning (SQIL) \citep{reddy2019sqil} also uses demonstrated transitions for both updates. However, in SQIL, in order to encourage the agent to imitate the demonstrated behavior, the demonstrated transitions are all associated with a reward of 1 while new transitions collected by the agent receive a reward of 0. 
Although both SACfD and SQIL are designed to work with many demonstrations (over $200$ in SACfD), we show that this simple mechanism is powerful enough to efficiently guide the policy learned by SR-DCIL.

Other approaches mixing RL with imitation-based adversarial approaches \citep{kang2018policy} or generative models \citep{wu2021shaping} have also been considered to perform RLfD, mostly to overcome exploration limits. 
However, even if these approaches significantly accelerate the underlying RL algorithms, none of them are designed to work with a single demonstration as they rely on often unstable generator-discriminator architectures \citep{dadashi2020primal}. 

\section{Background}
\label{sec:background_dcil}

\subsection{Goal-Conditioned Reinforcement Learning}

To formalize Goal-Conditioned RL (GCRL) problems \cite{Kaelbling1993learningto, Moore1999multi, schaul2015universal, nasiriany2019planning, chane2021goal}, one call upon Markov Decision Processes $\mathcal{M} = (\mathcal{S}, \mathcal{A}, R, p, \gamma)$ and extend it with a goal space $\mathcal{G}$. 
At each step $t=0,1,2,...,T$ of a rollout, the agent is aiming for a goal $g_{t}$  and selects an action $a_{t}\in \mathcal{A}$ based on its current state $s_{t}\in \mathcal{S}$ using a policy $\pi(a_{t}|s_{t}, g_{t})$. 
It then moves to a new state $s_{t+1}$ according to an unknown transition probability $p(s_{t+1}|s_{t},a_{t})$ and receives a reward via an unknown reward function $R:\mathcal{S} \times \mathcal{A} \times \mathcal{S} \times \mathcal{G} \rightarrow \R$. A discount factor $\gamma$ which may depend on $g_{t}$ \cite{schaul2015universal} describes the importance of long-term rewards in this sequence of interactions \cite{sutton1998introduction}. 

The objective in GCRL is to obtain a Goal-Conditioned Policy (GCP) that maximizes the expected cumulative reward. 

\begin{equation}
\label{eq:cumulative_reward_RL}
\mathop{\mathbb{E}}_{\pi}[\sum_{t}\gamma^{t} R(s_{t}, a_{t}, s_{t+1}, g_{t})].
\end{equation}

\subsection{Distance-based sparse reward}
\label{sec:distance_based_reward}

Usually, in GCRL, the reward function is sparse: agents receive a reward of 1 if the goal is reached, 0 otherwise. 
However, in a complex, high-dimensional state space, achieving a precise desired state can be prohibitively expensive. 
Therefore, goals usually correspond to low-dimensional projections of states \citep{nachum2018data, ecoffet2021first, bagaria2021robustly} and the agent must reach any state whose projection is close enough (often according to a L2-norm) to the desired goal. These states constitute the success states set $S_{g}$. Thus, a goal-conditioned reward function is generally defined as follows:

\begin{equation}
\label{eq:distance_reward}
    R(s,g) = \begin{cases}
1 \text{ if } s \in \mathcal{S}_{g} \\ 
0 \text{ otherwise.}
\end{cases}
\end{equation}



\subsection{The DCIL framework}
\label{sec:bg_dcil}

In DCIL, the agent must reach a distant goal $g$ in the absence of any informative reward function.
To avoid a potentially long exploration process \citep{ecoffet2019go, ecoffet2021first}, a demonstration containing a sequence of states leading to $g$ is provided. 
From this demonstration, a sequence of goals $\tau_{\mathcal{G}} = (g_{1}, g_{2}, ..., g_{N-1}, g)$ is extracted and used to guide the learning process. 
The objective is to learn a GCP to guide the agent through this sequence of goals.

In this context of sparse rewards, the naive approach is to use a distance-based sparse reward to learn this policy. 
However, there is a risk that the agent will learn a sub-optimal policy and reach intermediate goals by reaching invalid success states i.e. success states from which the agent cannot reach the next goal, making it impossible to reach the goal sequentially.

The DCIL algorithms \citep{chenu2022leveraging, Chenu2022divide} are specifically designed to force the agent to learn how to reach each intermediate goal by reaching only valid success states.
Indeed, they increase the value of valid success states to ensure that their value is greater than the value of all other states. 
To achieve this, in DCIL-II, in the event of success ($s_{t} \in \mathcal{S}_{g_{t}}$), the current goal is automatically replaced by the next one in the sequence. 

Therefore, using the Bellman equation, the value of a success state is approximated as follows when $\gamma = 1$:

\begin{equation}
\label{eq:value_propagation_V_prep}
\begin{split}
V(s_{t}, g_{i}) &= R(s_{t},a_{t}, s_{t+1},g_{i}) + \gamma V(s_{t+1}, g_{i+1}) \\
&= 1 + V(s_{t+1}, g_{i+1})
\end{split}
\end{equation}

As a result, the value $V(s_{t+1}, g_{t+1})$ is propagated into $V(., g_{t})$. 
If V is well approximated, $V(s_{t+1}, g_{i+1})$ is larger for valid success states, the agent is encouraged to reach only valid success states and the following holds: 

\begin{equation}
\label{eq:value_propagation_V}
V(s_{t}, g_{i}) = \begin{cases}
2 \text{ if } s_{t+1} \text{ is valid}, \\
1 \text{ otherwise.}
\end{cases}
\end{equation}

Correctly estimating the value of success states can be challenging. DCIL-II relies on two key principles to obtain a good estimate of the value: it exploits a strong reset assumption to simplify this estimation (see Section~\ref{sec:methods_sr_dcil}) and it relies on the Hindsight Experience Replay (HER) relabeling mechanism \cite{andrychowicz2017hindsight} which is crucial for efficient learning in a sparse reward context. 
Note that in order to make the value propagation mechanism compatible HER relabeling mechanism, the agent must keep track of the current target goal. In DCIL-II, this is achieved by augmenting states with the index of the current intermediate goal in the sequence. 

\section{Methods}
\label{sec:methods_sr_dcil}

In this section, we start by highlighting the importance of a strong reset hypothesis in the GCRL framework of DCIL-II. Then, we present two mechanisms, the Demo-Buffer and Value Cloning, designed to encourage the agent to reach valid success states only. In addition, we present Approximate Goal Switching, a mechanism designed to help the agent to train more efficiently with this weaker reset assumption. Finally, we present the complete algorithm called Single-Reset DCIL (SR-DCIL).

\subsection{The importance of the reset hypothesis}
\label{sec:important_reset_hypothesis}

SR-DCIL is couched in the DCIL framework. 
In this framework a sequence of goals $\tau_{\mathcal{G}}=\{g_{i}\}_{[1,N_{goal}]}$ is extracted from a state-based demonstration and used to guide the agent to a difficult to achieve goal.

\begin{figure}[!ht]
     \centering
     \includegraphics[width=\hsize]{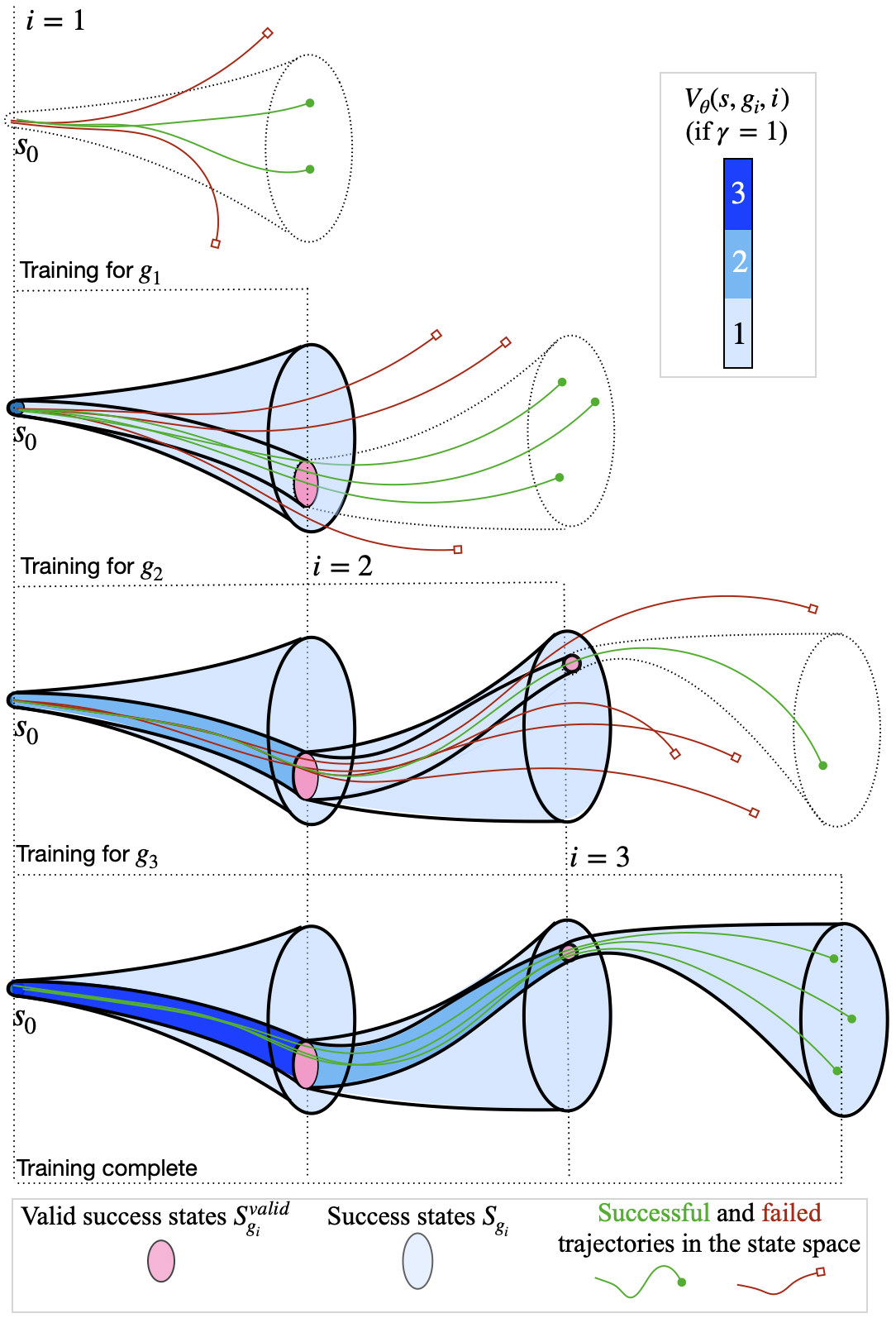}
    \caption{Limitations caused by a reset to a single state. The agent needs to reach $g_{1}$ to train for $g_{2}$ and to reach both $g_{1}$ and $g_{2}$ to train for $g_{3}$. Moreover, to successively reach each goal, the agent must transit between successive sets of valid success states (in pink). Until the agent learns how to reach the second goal, the values of valid and invalid success states associated with the first goal are similar ($V_{\theta}(s,g_{1},1) \approx  1, \forall s \in \mathcal{S}_{g_{1}}$ until the agent learned how to reach $g_{2}$). Therefore, the agent is not encouraged to target valid success states. This results in a large number of wasted training trajectories (in red), as they were launched from invalid success states. When the agent mostly reaches invalid success states (as for goal $g_{2}$ here, which has a small set of valid success states), training for the next goal can become very challenging as most training rollouts for $g_{3}$ start from incompatible states.}
    \label{fig:value_valid_success_states}
\end{figure}

DCIL-II learns a GCP to reach each goal in $\tau_{\mathcal{G}}$ and uses a value propagation mechanism to encourage the agent to achieve each goal by reaching valid success states for the following goals in the sequence. 
However, in DCIL-II, it is assumed that the agent can be reset to demonstrated states to learn how to reach each goal individually. 
If the reset assumption is weaker and the agent can only be reset to a single state, two limits arise. First, it gets difficult to train for further goals in the sequence. Moreover, the value propagation mechanism is not enough to encourage the agent to reach goals via valid success states. Those two limits are illustrated in Figure~\ref{fig:value_valid_success_states}.

\subsubsection{Training for distant goals}

In the GCRL framework of DCIL-II, the goal targeted by the agent is switched to the next in $\tau_{\mathcal{G}}$ and the index is incremented only if the current goal has been achieved. 
If the agent is only reset to a single state, the agent must be able to sequentially reach the $i-1$ previous goals in order to begin a training rollout for goal $g_{i}\in \tau_{\mathcal{G}}$. 
Therefore, the further the goal is in the sequence, the fewer training rollouts are performed by the agent for this goal. 

Moreover, the use of a stochastic policy makes narrow goal misses inevitable. 
In the context of non-holonomic environment, an unrecoverable miss of the current goal $g_{i}$ is likely to occur. After missing the goal, the agent continues to aim for $g_{i}$ until it is reset to its initial state, either after a time limit or after reaching a terminal state. 
In the meantime, no training transition for goal $g_{i+1}$ has been collected. 

On the contrary, in DCIL-II, to train for this same goal $g_{i}\in \tau_{\mathcal{G}}$, the agent is reset to a demonstrated state taken few steps ahead along the demonstration.
Thus, the agent needs very little exploration to learn how to reach $g_{i}$ as the number of control steps to reach goal $g_{i}$ from the associated reset state is limited. After enough training, the agent should be able to reach any goal in $\tau_{\mathcal{G}}$ starting from the associated demonstrated reset states. 

\begin{figure*}[h!]
     \centering
     \includegraphics[width=\hsize]{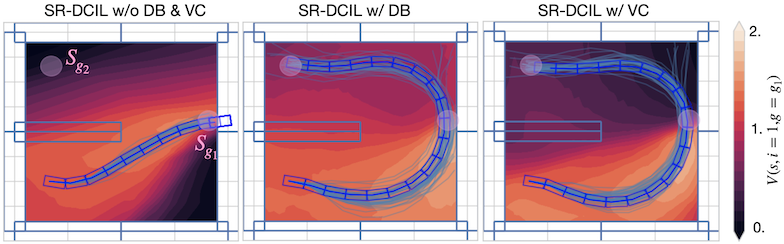}
    \caption{Visualising the impact of the Demo Buffer and Value Cloning after 15k training steps. When SR-DCIL is not equipped with a DB or VC (SR-DCIL w/o DB \& VC), the agent is not guided to valid success states by an increase of the Q-value of demonstrated state-action pairs (as in SR-DCIL w/ DB) or an increase of the value of demonstrated states (as in SR-DCIL w/ VC). As a result, it fails to achieve $g_{1}$ via valid success states and cannot reach $g_{2}$. On the contrary, SR-DCIL w/ DB and SR-DCIL w/ VC encourage the agent to achieve $g_{1}$ by reaching valid success states and manage to reach $g_{2}$.}
    \label{fig:vc_db_visu}
\end{figure*}

\subsubsection{Propagating the value between successive goals}

If a single reset state is available, as long as the agent has not learned how to reach goal $g_{i}$, the value of the valid and invalid success states associated with $g_{i-1}$ should be similar.
Indeed, both types of success states have a value close to one as the reward received for reaching the next goal has not yet been discovered and included in the value of valid success states. 
Therefore, the agent is not encouraged to achieve goal $g_{i-1}$ by reaching valid success states. This can prevent successful training. 

For instance, if the agent comes across valid success states associated with $g_{i-1}$ often enough by chance, after sufficient training for $g_{i}$, the reward received when reaching it increases the value of these valid states and the agent is encouraged to reach them. 
However, if the agent only comes across invalid success states, it is never able to reach $g_{i}$ nor to propagate the value containing the associated reward. 
Thus the agent does not distinguish valid and invalid success states.

On the contrary, in DCIL-II, when the agent is reset to a demonstrated state to train for goal $g_{i}$, the demonstrated state corresponds to a valid success states for previous goal $g_{i-1}$.
After enough training (i.e. successful training trajectories triggering the distance-based reward), this demonstrated valid success state should have a higher value than invalid states. Therefore, by propagating the value between the two successive goals via \eqref{eq:value_propagation_V}, the agent is encouraged to target this demonstrated valid success state as its value is higher than any other success state. 

\subsection{Using DCIL in the single-reset setting}

To adapt DCIL-II to the single-reset setting, we integrate different mechanisms. 
Depending on whether the expert actions are available, a \textit{Demo-Buffer} or a \textit{Value Cloning} mechanism can increase the value of valid success states. 
In addition, \textit{Approximated Goal Switching} (AGS) can help the agent training for distant goals.

\subsubsection{Increasing the value of valid success states}

In this section, we present two mechanisms: the Demo-Buffer (DB) and the Value Cloning (VC), designed to augment the value of the valid success states.

The DB uses the demonstration transitions when learning the value function. It therefore increases the Q-value of demonstrated state-action pairs, which necessarily pass through valid success states.
In the DB mechanism, we assume that expert actions are available, which is an additional requirement compared to the state-based demonstration necessary in DCIL-II. 

On the contrary, VC is an alternative to the DB mechanism if only a state-based demonstration is available.  
The VC calculates a theoretical value for each state of the demonstration and forces the value function of the agent to respect these theoretical values.
The agent then identifies a path in the state space to the valid success states of the demonstration. 

\paragraph{Demo-buffer}
\label{sec:DB}
The Demo-Buffer (DB) is a secondary Replay Buffer (RB) used during the actor and critic updates of SAC in the same way as SACfD \citep{martin2021learning, vecerik2017leveraging, paine2019making}. 
While the usual RB collects the training transitions, the DB is filled with the transitions extracted from the demonstration. 
During each SAC update, a batch of transitions is partly sampled from both buffers. The batch is filled with $80\%$ of training transitions and $20\%$ of transitions extracted from the demonstration. 
This $80-20$ ratio constitutes an additional hyper-parameter and has been chosen empirically. 

The demonstration trajectory contains a successful transition to each goal leading to the demonstrated valid success state. 
Therefore, by updating the critic networks using the Mean Squared Bellman Error (MSBE) \eqref{eq:msbe} for these demonstrated transitions, the value of the state-action pairs leading to the demonstrated valid success states is increased (see Figure~\ref{fig:vc_db_visu}). 

\begin{equation}
\begin{split}
\label{eq:msbe}
&\mathcal{L}(\theta, \mathcal{D}_{train},\mathcal{D}_{demo}) = \mathop{\mathbb{E}}_{(s,i,g,a,r,s',i',g')\sim \mathcal{D}_{train} \cup \mathcal{D}_{demo}}\\
&\begin{bmatrix}\frac{1}{2}
\begin{bmatrix}
Q_{\theta}(s,i,g,a) - (r + \gamma(s') Q_{\theta}(s',i',g',a))
\end{bmatrix}^2
\end{bmatrix}.
\end{split}
\end{equation}

Thus, when reaching goal $g_{i-1}$ in $\tau_{\mathcal{G}}$, the agent is encouraged to target the associated demonstrated valid success state. This prevents the agent from training for next goal $g_{i}$ by starting from an invalid success state.

However, in order to use the DB mechanism, we make a new assumption that was not necessary in DCIL-II: the internal actions of the demonstration should be available to fill the DB.

\begin{figure*}[h]
     \centering
     \includegraphics[width=0.9\hsize]{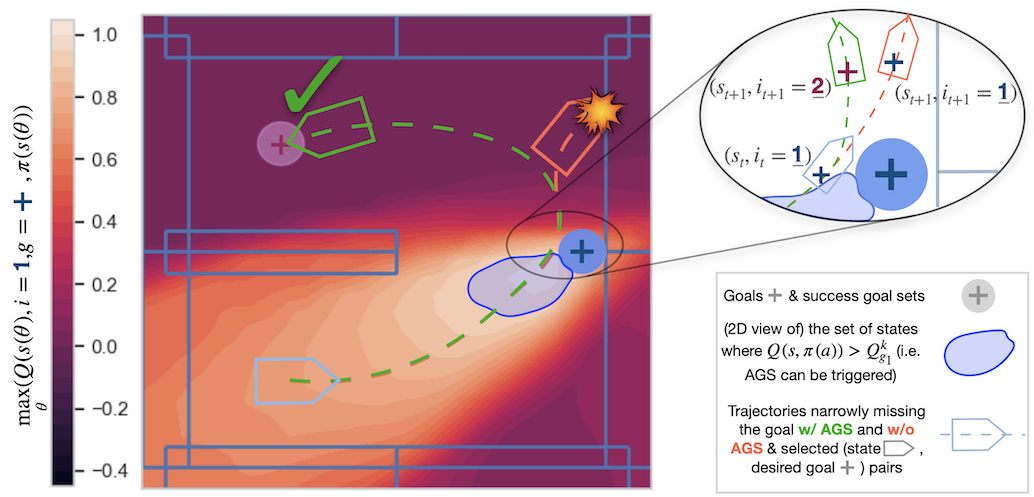}
    \caption{Illustration of the Approximated Goal Switching concept in a toy 2D maze where the agent corresponds to a Dubins Car \citep{dubins1957curves} with $(x,y,\theta)$ states and $(x,y)$ goals. The contours represents the maximum value function obtained in the $(x,y)$ position by uniformly sampling 20 orientations, after 15k SR-DCIL training steps. In the green trajectory, the agent triggered AGS by entering the blue zone. $k$-steps after, the goal is automatically switched to the next one in $\tau_{\mathcal{G}}$, and the agent can continue its progression in the maze. On the contrary, in the red trajectory w/o AGS, after the irrecoverable narrow miss of the first goal, the agent is still conditioned on this goal and eventually collides with the wall while trying to turn toward the goal (for better readability, we drop the index and the goal in the Q-function entry when it is not necessary).}
    \label{fig:ags_concept}
\end{figure*}

\paragraph{Value cloning}
\label{sec:VC}
In Value Cloning (VC), we only assume access to the states of the demonstration. 
Each state of the demonstration is adapted to the DCIL framework, by adding the associated objective and index in $\tau_{\mathcal{G}}$. 
This associated goal corresponds to the closest goal in $\tau_{\mathcal{G}}$ extracted from a state further along the demonstration. 

Using the demonstrated trajectory, we can compute for each state the theoretical return $V_{demo}$ received by the agent by passing through the remaining demonstrated states. 
This theoretical return is computed according to the distance-based reward, the following transition and the discounted functions defined by the DCIL framework.
It corresponds to the discounted sum of sparse rewards received for reaching each goal sequentially. 
The set of demonstrated states and their associated values form the Value Cloning dataset $\mathcal{D}_{VC}$.

During each SAC update, a batch of states is sampled partly from the training transitions and partly from $\mathcal{D}_{VC}$. The batch is filled with $80\%$ of states coming from the training RB and $20\%$ coming from $\mathcal{D}_{VC}$ similarly to DB. 
For states coming from the training RB, their associated target value corresponds to the on-policy soft Q-value computed with respect to the Q-value critic network $Q_{\omega}$ \cite{haarnoja2018soft}. This corresponds to a usual SAC update.
For states extracted from $\mathcal{D}_{VC}$, the associated value corresponds to the theoretical one computed according to the demonstration.

The demonstrated states indicate a path towards valid success states. Therefore, by updating the value critic using the Mean Squared Error (MSE) between the current estimate of the value of demonstrated states and their theoretical value, their value is increased and the agent is encouraged to pass through these demonstrated states while reaching the goals in $\tau_{\mathcal{G}}$ (see Figure~\ref{fig:vc_db_visu}).

\begin{equation}
\begin{split}
\label{eq:mse_vc}
&\mathcal{L}_{VC}(\theta, \mathcal{D}_{train}, \mathcal{D}_{VC}) = \\ &\mathop{\mathbb{E}}_{(s,i,g,V_{target})\sim \mathcal{D}_{train}\cup\mathcal{D}_{VC}}
\begin{bmatrix}\frac{1}{2}
\begin{bmatrix}
V_{\psi}(s,i,g) - V_{target}
\end{bmatrix}^2
\end{bmatrix},
\end{split}
\end{equation}

with 

\begin{equation}
V_{target} = 
\begin{cases}
Q(s,\pi(s),i,g) \text{ if } (s,i,g)\in \mathcal{D}_{train}\\
V_{demo} \text{ if } (s,i,g)\in \mathcal{D}_{VC} 
\end{cases}
\end{equation}

One should note that this mechanism requires the original Actor-Critic architecture of SAC where the critic contains two networks estimating the value function and the Q-value function. 
In recent implementations of SAC, only the Q-value critic remains \citep{haarnoja2018softapp}. 

\subsubsection{Increasing the number of rollouts for distant goals}

To avoid wasting a training rollout where the agent narrowly missed the goal but still managed to reach a state from which the next goal is achievable, Approximated Goal Switching (AGS) changes the current goal automatically whether the agent reaches it or misses it.

As soon as the agent reaches a state close enough to the goal so that its Q-value is large enough, whether the agent actually reaches it or narrowly misses it a few step later, the index is automatically incremented and the current goal is switched for the next one in $\tau_{\mathcal{G}}$. 
Therefore, the agent starts training for the next goal in $\tau_{\mathcal{G}}$.

For each goal $g_{i} \in \tau_{\mathcal{G}}$, we consider that the Q-value of a state close to $g_{i}$ is large enough if it is above a threshold $Q^{k}_{g_{i}}$. This threshold corresponds to the maximum Q-value associated with a state taken $k$-steps ahead of the success state along a successful training rollout reaching $g_{i}$. This threshold is illustrated in Figure~\ref{fig:ags_concept}. In each experiment of this paper, $k$ is a hyper-parameter set to $2$. 

During the remaining $k$ steps, the agent has enough control steps left to reach the goal. Therefore, the successful transition required to propagate the value from the next goal to the previous one in the GCRL framework of DCIL-II (see \eqref{eq:value_propagation_V}) is still collected. Moreover, if the agent narrowly misses the goal, it can still perform a training rollout for the next goal.

The AGS mechanism avoids premature termination of training rollouts when the agent has narrowly missed a target. Therefore, it increases the number of training rollouts performed for remote goals in $\tau_{\mathcal{G}}$.

\begin{algorithm}[!ht]
\caption{SR-DCIL}
\label{algo:RF_DCIL}
\begin{algorithmic}[1]
    \State \textbf{Input:} $\pi, Q, \bar{Q}, \tau_{demo}$ 
    \Comment{actor/critic/target critic networks \& demonstration}
    \State $\{g_{i}\}_{i\in[1,N_{goals}]},$ $\mathcal{D}_{demo/VC}\leftarrow$ extract$(\tau_{demo})$ \color{black}
    \State $B\leftarrow [\ ]$ 
    \Comment{replay-buffer}
    \State $Q_{max} = \{0\}_{i\in[1,N_{goals}]}$ \color{black}
    \Comment{Q-value thresholds (AGS)}
    
    \For{$n=1:N_{episode}$}
    \Comment{\textbf{Trajectory initialization}}
        \State $ i, T_{left},  s \leftarrow 0, T_{max}, \text{env.reset}()$
        \State $success, done, last\_index \leftarrow False, False, False$ \label{lst:line:end_traj_init2}
        \State  $D_{queue}, b_{AGS} \leftarrow [],$ False \color{black}
        \Comment{AGS memory + boolean}
        \While{not done} 
        \Comment{\textbf{Trajectory rollout}} \label{lst:line:begin_traj_rollout2}
            \State $a \sim \pi(a|s,i,g_{i})$
            \State  $D_{queue}.insert((s,a,i))$ \color{black}
            \State $s', env\_done, r \leftarrow env.step(a), 0$ 
            \If{$s' \in \mathcal{S}_{g_{i}}$}
            \Comment{success}
            \label{lst:line:success2}
                \State $r, success \leftarrow 1, True$
                \State $last\_index \leftarrow (i \geq nb\_skills)$
                \State $i' \leftarrow i+1$
                \Comment{index shift}
                \State  $Q_{max} \leftarrow update\_threshold (Q_{max}, D_{queue}, \tau_{\mathcal{G}})$ \color{black}
                \label{lst:line:ags_update2}
            \ElsIf{$\bar{Q}(s,i,g_{i},a) \geq Q_{max}$\color{black}}
                \State $T_{left} \leftarrow k$\color{black}
                \Comment{AGS activation}
                \label{lst:line:ags_activation2}
                \State $b_{AGS} \leftarrow$ True\color{black}
            \ElsIf{$T_{left} \leq 0\ \text{or}\ env\_done$}
                \Comment{failure}
                \If{$b_{AGS} = $ True\color{black}} 
                    \Comment{AGS switch}
                    \label{lst:line:ags_switch2}
                    \State  $success, timeout \leftarrow False, False$ \color{black}
                    \State $i' \leftarrow i+1$\color{black}
                \Else
                    \State $success, timeout \leftarrow False, True$
                    \State $i' \leftarrow 0$
                \EndIf
            \Else
                \State $success, timeout \leftarrow False, False$
                \State $i' \leftarrow i$
            \EndIf
            \If{$env\_done\ \text{or}\ last\_index$}
                \State $done \leftarrow True$
            \label{lst:line:tack_failure2}
            \EndIf
            \State $T_{left} \leftarrow T_{left} -1$
            \State $B \leftarrow B + (s, g_{i}, i, a, s', g_{i'}, i', r, done, success)$
            \State $s,i \leftarrow s',i'$
            \State $done \leftarrow done \lor timeout$ \label{lst:line:end_traj_rollout2}
            \State $\text{SAC\_update}(\pi, Q, \bar{Q},B, \mathcal{D}_{demo/VC})$ \label{lst:line:sac_update2}
        \EndWhile
    \EndFor
\end{algorithmic}
\end{algorithm}

\subsection{The SR-DCIL algorithm}
\label{sec:rf_dcil_algo}

Given a single demonstration, the SR-DCIL algorithm first extracts the sequence of goals and the elements required to construct the Demo-Buffer or to perform Value Cloning. The DCIL GC-MDP is derived by extending states with goals and indices exactly as in DCIL-II \cite{chenu2022leveraging}. 
Then, SR-DCIL runs a 2-step loop to learn a policy that can be used to reach each goal sequentially. By doing so, the agent is able to complete the complex demonstrated behavior. Algorithm \ref{algo:RF_DCIL} summarizes these different steps.

\subsubsection{Processing the demonstration}

The sequence of goals is extracted in the same way as in DCIL-I and DCIL-II.
The demonstrated states are projected in the goal space and the demonstration is split into $N_{goal}$ sub-trajectories of equal arc lengths $\epsilon_{\text{dist}}$. For each sub-trajectory in the goal space, we extract its final elements and concatenate them to construct $\tau_{\mathcal{G}}$.

If the internal actions of the demonstration are available, the DB can be constructed using all the (state, action, next state) transitions of the demonstration. The states and next states in each transition are augmented with their associated goal in $\tau_{\mathcal{G}}$ and the corresponding index.

On the contrary, if internal actions are not available, we rely on VC to increase the value of valid success states.  
In that case, we construct the VC dataset $\mathcal{D}_{VC}$ by collecting all the states in the demonstration, augmenting them with their associated goal and index and calculating their theoretical discounted return. 

\subsubsection{Main loop}

SR-DCIL repeatedly performs trajectory rollouts in the environment using Approximated Goal Switching to collect training transitions. 
It combines an off-policy actor-critic algorithm (e.g. the Soft Actor-Critic (SAC) algorithm \citep{haarnoja2018soft}) with the HER-like relabelling mechanism of DCIL-II and DB or VC to learn the goal-conditioned policy. 

\paragraph{Collecting transitions}

SR-DCIL resets the agent in the unique reset state at the beginning of the demonstration. The policy is conditioned on index $1$ and the first goal $g_{1}$ in $\tau_{\mathcal{G}}$. The agent then starts a trajectory.
During this trajectory, the current goal is switched to the next one in $\tau_{\mathcal{G}}$ if the agent achieves this current goal (line~\ref{lst:line:success2}) or if AGS is triggered (lines~\ref{lst:line:ags_activation2} and \ref{lst:line:ags_switch2}). 

If the agent actually reaches the current goal, the associated Q-value threshold $Q_{g_{i}}^{k}$ used to trigger AGS is potentially updated (line~\ref{lst:line:ags_update2}): 
First, the Q value of the state-action pair taken k steps before success is calculated. Then, if this value is greater than the current threshold, it replaces it. 

If the agent reaches each goal successively up to the final one, if it reaches a terminal state or if a time limit is reached the current trajectory is interrupted and the agent is reset to the unique reset state. 

\paragraph{Policy update}

SR-DCIL performs a SAC update after each step in the environment. 

If we use a DB to increase the value of valid success states, 80\% of the sampled batch of transitions used to perform the actor-critic update are training transitions and 20\% are demonstrated transitions from the DB.  
In the sampled batch of transitions, half of the transitions are relabeled using the relabelling mechanism of DCIL-II \citep{chenu2022leveraging}. 

If we use VC instead of DB, 20\% of the training batch for the value network updates contains demonstrated states. Their target values correspond to the theoretical value (see Section~\ref{sec:VC}). The update of the actor remains unchanged.


\section{Experiments}
\label{sec:SR_DCIL_xp}

In this Section, we start by presenting the experimental setup. 
We then present an ablation study of the mechanisms designed to increase the value of valid success states and to facilitate training for advanced goals in the sequence. 
Finally, we compare our method to DCIL-II to assess the loss of sample efficiency induced by a weaker reset assumption. 

\subsection{Experimental setup}

We evaluate SR-DCIL in three environments: the \textit{Dubins Maze} environment \citep{Chenu2022divide} and the \textit{Fetch} environment \citep{ecoffet2021first}. 

\subsubsection{Dubins Maze}

The \textit{Dubins Maze} is a navigation task where the agent controls a Dubins car \cite{dubins1957curves} in a 2D maze. The state $s=(x,y,\theta) \in X\times Y\times \Theta$ includes the 2D position of the car in the maze and its orientation. The forward velocity being constant, the agent only controls the variation of orientation $\dot{\theta}\in \mathbb{R}$ of the car. The goal space is defined as $X\times Y$, thus goals correspond to 2D positions. Such goal space design does not condition the orientation of the car when the agent reaches a goal. Demonstrations are obtained using the Rapidly-Exploring Random Trees (RRT) algorithm \cite{lavalle1998rapidly}.    

\subsubsection{Fetch}

The \textit{Fetch} environment is a simulated grasping task for a 8 degrees-of-freedom robot manipulator. A sparse reward is obtained only when the agent grasped an object and put it on a shelf. 
The state $s\in \mathbb{R}^{604}$ contains the Cartesian and angular positions and the velocity of each
element in the environment (robot, object, shelf, doors...) as well as the contact Boolean evaluated for each pair of elements. 
In this environment, a goal $g\in \mathbb{R}^{6}$ corresponds to the concatenation of the Cartesian position of the end-effector of the robot and the object. Therefore, the agent may reach a goal with an invalid orientation or velocity that may prevent grasping. 
Demonstrations are obtained using the exploration phase of the Go-Explore algorithm \cite{ecoffet2021first}.

\subsection{Baseline}

To evaluate the drop in efficiency induced by resetting the agent to a single state, we compare SR-DCIL to DCIL-II. 
Indeed, by resetting the agent to demonstrated states, DCIL-II not only overcomes the limits underlined in Section~\ref{sec:important_reset_hypothesis}, but it also learns a complex behavior by training on short rollouts only \cite{chenu2022leveraging}. 
Therefore, DCIL-II should be more sample efficient than RF-DCIL at learning complex behaviors. 

\subsection{Ablation study}
\label{sec:ablation_SR_DCIL}

Using the Dubins Maze and the Fetch environments, we compare five variants of SR-DCIL. 
Two variants benefit from the availability of demonstrated action and use DB to increase the value of valid success states. 
Two others assume that demonstrated actions are not available and use VC instead. 
In both cases, one variant (called \textbf{DB w/ AGS} or \textbf{VC w/ AGS}) uses AGS which helps the agent train for the furthest goals in the sequence. 
The others (called \textbf{DB w/o AGS} or \textbf{VC w/o AGS}) do not use AGS.
A final variant called \textbf{vanilla} corresponds to the application of DCIL-II in the context of a reset to a single state. This variant does not benefit from any mechanism to guide the agent toward valid success state and does not use AGS to help the agent train on distant goals. 

While the \textit{Dubins Maze} validates the utility of each mechanism, conducting a similar ablation study in \textit{Fetch} is mandatory to evaluate the impact of high-dimensional states and action spaces on them.

\begin{figure}[ht]
     \centering
     \includegraphics[width=\hsize]{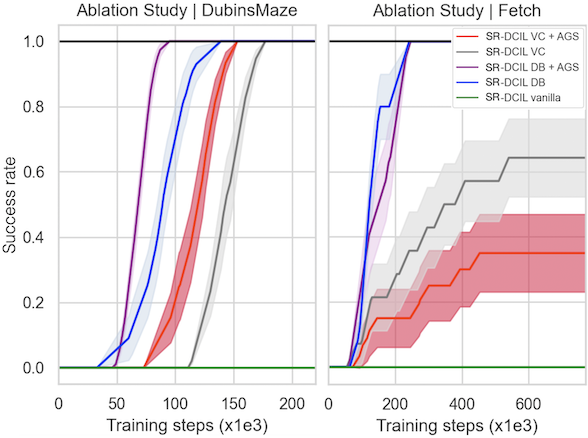}
    \caption{Ablation study. Comparing different variants of SR-DCIL in the Dubins Maze and the Fetch environments: we evaluate the success rates of SR-DCIL with two different mechanisms (EB and VC) to encourage the agent to reach each goal via valid success states. Both variants are evaluated with and without AGS. In addition, we evaluate a vanilla version of SR-DCIL without AGS, VC or DB equivalent with DCIL-II with a reset to a single state. The mean and standard deviation are computed over 10 seeds. The standard deviation is divided by two for better visualization.}
    \label{fig:ablation_RF_DCIL}
\end{figure}

Figure~\ref{fig:ablation_RF_DCIL} presents the proportion of runs that solved the maze depending on the number of training steps. 
First, we can notice that in both environments, the variants of SR-DCIL using the DB benefit from the additional information contained in the demonstration and outperform the variants using VC. 

In addition, we can notice that the AGS mechanism results in a significant gain of performance in the Dubins Maze both for SR-DCIL VC and SR-DCIL DB. 
However, in the Fetch environment, SR-DCIL DB w/ AGS and SR-DCIL VC w/ AGS perform worse than their counterpart without AGS. 
We believe that two elements may be responsible for this difference in performance. 
On the one hand, the sequence of goals is shorter in Fetch compared to Dubins Maze (7 goals in Fetch compared to 17 goals in Dubins Maze). Therefore, it is easier to train for every goal in the sequence in Fetch than in Dubins Maze. 
On the other hand, while the Q-value threshold generalizes well in the low-dimensional state space of the Dubins Maze (see its intuitive form in \figurename~\ref{fig:ags_concept}), it is difficult to ensure that it has an analogous form in the high-dimensional state space of Fetch. In particular, poor generalization may easily occur, resulting in an unexploitable threshold.

\begin{figure}[ht]
     \centering
     \includegraphics[width=0.9\hsize]{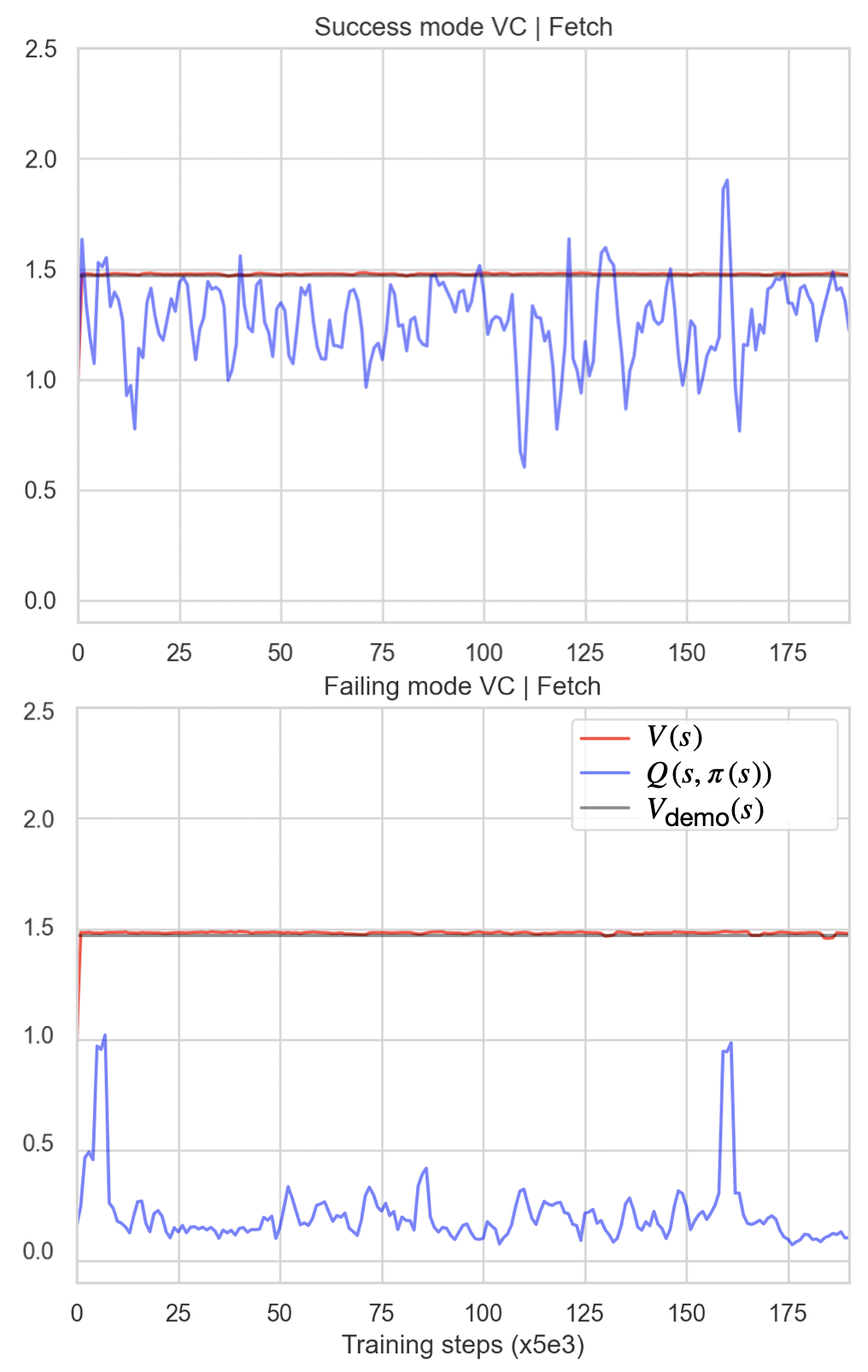}
    \caption{Success and failure modes of VC. In the Fetch environment, $35\%$ of the runs fail to learn how to grasp the object. As expected, the Value Cloning mechanism sets the value of the demonstrated states to their theoretical value. Here, in the two selected runs, the evolution of the learned value $V(s)$ of the last demonstrated state before the complex grasping behavior is plot in red and matches with the theoretical value in black. In the successful run (top panel), the agent visits states similar to the demonstrated states. Therefore, the high value is propagated in the Q-value (on-policy Q-value of the last state before grasping in blue) which impacts the policy and guides the agent toward valid success states.  However, in the failing run (bottom panel), those states are hardly visited by the agent while training. Therefore, their high value has little to no impact on the Q-value (on-policy Q-value of the last state before grasping in blue) and the policy. As a result, the agent is not guided toward valid success states.}
    \label{fig:failing_mode_VC}
\end{figure}

Finally, one should notice that, in Fetch, SR-DCIL VC only reaches a $65\%$ success rate as, in $35\%$ of the runs, the agent fails to achieve the first goals via valid success states and to learn how to grasp the object.
Indeed, as shown in Figure~\ref{fig:failing_mode_VC}, even if the Value Cloning mechanism artificially increases the value of the demonstrated states as expected, during these failed runs, the cloned value of demonstrated states has no impact on the Q-value and the policy. Thus, the agent is not guided toward valid success states. 

We believe that this absence of impact of the cloned value of demonstrated states on the Q-value results from the fact that the agent never visits states close to the demonstrated ones. 
Indeed, if no transition to these states is collected in the training RB, no update of the Q-value involving the cloned value can be performed. The agent is never encouraged to navigate the demonstrated states while reaching the goals.

One might think that increasing the entropy coefficient of SAC would be sufficient to encourage the agent to explore more and eventually find the demonstrated states. 
However, according to our observations, increasing this coefficient does not prevent these failure modes from appearing as it does not allow the agent to extensively explore the state space. 
For instance, the demonstrated states correspond to particularly slow approach speeds. In contrast, the agent is encouraged by the discount factor to reach each goal as quickly as possible.
Thus, the optimal character of the training trajectories may prevent the agent from exploring states corresponding to slower speeds and, therefore, prevents the agent from exploring states close to those of the demonstration. 

\section*{Discussion \& Conclusion}
\label{sec:discussion}

In this paper, we have attempted to relax the agent reset assumption in demonstration states. 
First, we highlighted the consequences of a weaker reset assumption to a single initial state: an increased difficulty in training for distant goals and propagating value between successive goals. 
Secondly, we proposed several mechanisms to compensate for these difficulties: the Approximated Goal Switching to overcome the difficulty of training for distant goals, the Demo-Buffer and Value Cloning to facilitate the propagation of value. 
Based on these mechanisms, we presented several variants of Single-Reset Divide \& Conquer Imitation Learning (SR-DCIL), which we tested in two environments of different complexity: a low-dimensional non-holonomic navigation task and a high-dimensional robotic manipulation task. 
While the proposed mechanisms appear to be effective in the navigation task, their performance is mixed in the manipulation task. 
We hypothesized that the performance of SR-DCIL decreases with increasing problem dimensionality. 
In our future work, we will continue this analysis in order to propose a scalable solution. 
An interesting approach might be to combine the demo buffer with a learned inverse model to retrieve demo actions if only states are available.

\section*{Acknowledgements}
This work was partially supported by the French National Research Agency (ANR), Project ANR-18-CE33-0005 HUSKI and was performed using HPC resources from GENCI-IDRIS (Grant 2022-A0111013011).



%
\bibliographystyle{IEEEtran}

%





\end{document}